\pdfoutput=1

\documentclass[11pt]{article}
\usepackage[]{acl}

\usepackage{times}
\usepackage{latexsym}
\usepackage[T1]{fontenc}
\usepackage[utf8]{inputenc}
\usepackage{microtype}
\usepackage{inconsolata}
\usepackage{booktabs}
\usepackage{graphicx}

\newcommand{\systemname}{\textsc{MunTTS}}
\newcommand*\samethanks[1][\value{footnote}]{\footnotemark[#1]}

\title{\systemname : A Text-to-Speech System for Mundari}

\author{\parbox{0.9\linewidth}{\centering{Varun Gumma\textsuperscript{\normalfont $\spadesuit$} \quad Rishav Hada\textsuperscript{\normalfont $\spadesuit$} \quad Aditya Yadavalli\textsuperscript{\normalfont $\diamondsuit$} \\ \quad Pamir Gogoi\textsuperscript{\normalfont $\heartsuit$}\thanks{Work done when the author was at Microsoft} \quad Ishani Mondal\textsuperscript{\normalfont $\clubsuit$}\samethanks \quad Vivek Seshadri\textsuperscript{\normalfont $\spadesuit\diamondsuit$} \quad Kalika Bali\textsuperscript{\normalfont $\spadesuit$}\\
 {\rm  \textsuperscript{\normalfont $\spadesuit$}Microsoft Corporation 
 \quad 
 \textsuperscript{\normalfont $\diamondsuit$}Karya Inc. 
 \quad
 \textsuperscript{\normalfont $\heartsuit$}Project VANI
 \quad 
 \textsuperscript{\normalfont $\clubsuit$}University of Maryland\\}
{\tt varun230999@gmail.com, kalikab@microsoft.com} }}}

\begin{document}
\maketitle
\begin{abstract}
We present \systemname, an end-to-end text-to-speech (TTS) system specifically for Mundari, a low-resource Indian language of the Austo-Asiatic family.  Our work addresses the gap in linguistic technology for underrepresented languages by collecting and processing data to build a speech synthesis system. We begin our study by gathering a substantial dataset of Mundari text and speech and train end-to-end speech models. We also delve into the methods used for training our models, ensuring they are efficient and effective despite the data constraints. We evaluate our system with native speakers and objective metrics, demonstrating its potential as a tool for preserving and promoting the Mundari language in the digital age\footnote{Artifacts available at \url{https://aka.ms/MUN-TTS}}.
\end{abstract}
\section{Introduction}
\label{sec:introduction}
India is home to approximately 1652 languages and 22 official languages written in different scripts \cite{9992058,gala2023indictrans}. Many of these languages are classified as low-resource, as native speakers are moving towards dominant languages that are supported by modern-day technologies \cite{bali-EtAl:2019:LT4ALL}.

According to a UNESCO report, India ranks fourth in the list of critically endangered languages with 41 languages \cite{kwan2022}. If a language becomes extinct we lose out a large part of the culture. Motivated by these factors, many communities have self-initiated data collection and partnered with tech organizations to build technologies for their language \cite{10.1145/3530190.3534792}.

Text-to-speech (TTS) systems have been gaining a lot of importance as a vital language technology due to their applications in education, navigation, accessibility, voice assistants etc. \cite{10096069}. However, the development of TTS systems for low-resource languages has several challenges \cite{pine-etal-2022-requirements}. Firstly, training current-day TTS systems requires many hours of audio recordings and corresponding text transcriptions which is resource-intensive. Secondly, carefully curating the training data such that it covers the phonetic complexity of the given language requires expert input. This becomes a problem especially when the available data is already scarce. Thirdly, the availability of native speakers who are familiar with technology and can do audio recordings in a high-quality studio setup. Fourthly, availability of enough native speakers, who could systematically evaluate these systems for subjective metrics. Lastly, getting high-quality audio recordings can be very expensive. Overcoming these challenges requires not only technical expertise to extract the most out of limited resources but also significant on-field operational efficiency to collect the right quality and quantity of data. No such data collection is possible without the active participation of the community and other stakeholders.

In this work, we build a TTS system for Mundari. Mundari is an Austro-Asiatic language spoken by Munda tribes in the eastern Indian states of Jharkhand, Odisha, and West Bengal. According to the 2011 India census, there are $\approx$1M native speakers of this language \cite{census2011}. Mundari is mainly written in the dominant script of the region where it is spoken, viz. Devanagari, Odia, and Bangla. In this study, we worked with the Mundari spoken in Jharkhand and written in the Devanagari script.

We collected audio recordings for 15,656 unique sentences in Mundari. Our Mundari speech corpus consists of high-quality 26,868 audio recordings in male and female voices consisting of 27.51 hours. The average duration of the recordings in our dataset is 3.7 seconds, and the average sentence length is 8.4 words. Using this data we train three TTS systems: Variational Inference with adversarial learning for end-to-end (E2E) Text-to-Speech (VITS) with 22KHz sampling rate (VITS-22K), VITS with 44KHz sampling rate (VITS-44K) and fine-tune XTTS v2 as well. We have audio samples from each of these systems evaluated by native speakers for subjective metrics. VITS-44K gives the best overall performance with MOS = 3.69 $\pm$ 1.18.

\section{Related Works}
\label{sec:related_works}
\begin{itemize}
    \item \textbf{Neural Speech Synthesis:} The field of neural speech synthesis has experienced a series of transformative developments. WaveNet \cite{oord2016wavenet}, utilized a convolutional neural network for raw audio waveform generation, marking a shift from previous heuristic synthesis methods. Tacotron \cite{wang17n_interspeech} further advanced the field with its end-to-end text-to-speech synthesis, streamlining the synthesis process. Tacotron 2 \cite{8461368} built upon this by incorporating WaveNet as a vocoder, enhancing speech quality. Parallel WaveGAN \cite{9053795} introduced a generative adversarial network approach for faster waveform generation. FastSpeech \cite{NEURIPS2019_f63f65b5} and FastSpeech 2 \cite{ren2021fastspeech} utilized feed-forward networks for faster speech generation and enhanced control over speech attributes. VITS \cite{pmlr-v139-kim21f} combined variational autoencoders \cite{9051780} with GANs \cite{NIPS2014_5ca3e9b1} in an end-to-end structure, enabling more expressive speech synthesis. Lastly, XTTS \cite{coqui_xtts} provided cross-lingual text-to-speech capabilities, representing a notable advancement towards adaptable speech synthesis systems. We refer the readers to \citet{tan2021survey} for a comprehensive survey of neural text-to-speech.
    
    \item \textbf{Relevant TTS systems:} In recent times, there has been a shift towards TTS models for low-resource languages, especially for Indian languages. \citet{vakyansh} put forth the first open-source monolingual neural systems for 9 Indic languages using a Glow-TTS + HiFi-GAN combination. \citet{prakash20_interspeech} advance it by releasing multilingual TTS models within the same family using a multilingual character map \cite{prakash19_ssw} and common label set \cite{prakash19_ssw} for Tacotron2 + WaveGlow. \citet{10096069} extend the language coverage to 13 by including 3 low-resource languages, Rajasthani, Bodo, and Manipuri. They also conduct a thorough analysis of different Non-Autoregressive (NAR), flow-based, and end-to-end models in a multi-speaker and multilingual setting and find that single-language models are preferable. Globally, \citet{pratap2023mms} expand the text-to-speech coverage to 1017 languages by training individual end-to-end VITS models for each language. However, none of them have developed a dedicated, high-quality, multi-speaker TTS for an extremely low-resource language. In this paper, we present our experiences in developing a high-quality, multi-speaker TTS model for such a language -- Mundari. 
\end{itemize}
\section{Data}
\label{sec:data}
Multiple steps were involved in the data collection process. First, the text data was obtained by translating a Hindi corpus of 100,000 sentences obtained from the Karya database. Karya\footnote{\url{https://karya.in/}} is a data services organization that takes requests from clients and breaks down these complex requests into simple microtasks that users with little to no digital literacy can perform.

We randomly selected 20,000 of these sentences and manually translated them to Mundari. The translated Mundari sentences were expressed using the Devanagari script. The translators were instructed to prefer fluency of the sentences over faithfulness of the translations wherever they had to make a choice. The translated sentences were then validated for appropriateness by native speakers. This text corpus was then used as the final dataset for recording one male and one female speaker. The male and female speaker was selected from a pool of 12 speakers (6 male and 6 female), who were asked to complete a reading task online. After they submitted the speech samples, the speakers were then evaluated by native speakers and given a score for their reading efficiency and pronunciation. Based on these scores, 3 male and 3 female speakers were shortlisted, and finally, from these 6 speakers, 1 male and 1 female speaker was selected after analyzing some voice quality features. The speakers, i.e., the voice artists, were instructed to record the sentences shown to them without any false starts, filled pauses, hiccups, or any other mistakes. All the recordings are done in a studio-quality room with a microphone connected to the Karya crowdsourcing application for the convenience of collecting the data. The recordings' sampling rate is 44.1 KHz with 32 bits per sample.

Finally, the curated text used to collect recordings contains 15,656 unique sentences. The average sentence length in the collected text corpus is 8.4 words. Some duplication of sentences across speakers yields a total of 26,868 sentences and recordings in our final dataset. Around 74\% of the recordings in our dataset feature a female speaker, while the remaining 26\% are attributed to a male speaker. We notice that the female recordings are, on average, slightly shorter than male recordings -- females' being 3.62 seconds compared to males' 3.85 seconds. We present more details in Table \ref{tab:data}. 

\begin{table}[t!]
\centering
\small
\begin{tabular}{@{}lccc@{}}
\toprule
\textbf{Data} & \textbf{Train} & \textbf{Val} & \textbf{Test} \\ \midrule
\textit{Avg. Sentence Length} & 8.48 & 8.57 & 8.44 \\
\textit{Total Duration (in hours)} & 24.76 & 1.379 & 1.375 \\
\midrule
\textbf{Male} \\
\midrule
\textit{Num of Recordings} & 6302 & 350 & 350 \\
\textit{Avg. Duration (in seconds)} & 3.85 & 3.80 & 3.88 \\
\textit{Total Duration (in hours)} & 6.74 & 0.37 & 0.38 \\
\midrule
\textbf{Female} \\
\midrule
\textit{Num of Recordings} & 17,879 & 993 & 994 \\
\textit{Avg. Duration (in seconds)} & 3.62 & 3.67 & 3.62 \\
\textit{Total Duration (in hours)} & 18.02 & 1.01 & 0.998 \\
\bottomrule
\end{tabular}
\caption{Dataset Metrics for the Mundari speech dataset.}
\label{tab:data}
\end{table}

\section{Experiments}
\label{sec:experiments}

\subsection{Pre-Processing}
\label{sec:preprocessing}
The source sentences were normalized by collapsing repeated punctuations, exclamations, and spaces. Next, all kinds of brackets were removed and newline and tab characters were substituted with spaces. The \texttt{indic\_nlp\_library}\footnote{\url{https://github.com/VarunGumma/indic_nlp_library}} was used to further normalize the Devanagari text and appropriately space words with ``matras''. The dataset was split into train (95\%), dev (5\%), and test (5\%) sets by stratifying on the number of speakers. The exact number of data points per split is available in Table \ref{tab:data}. 

\subsection{Models}
\label{sec:models}
We train E2E TTS models using the \texttt{coqui-ai}\footnote{\url{https://github.com/coqui-ai/TTS}} framework. These include a VITS model \cite{pmlr-v139-kim21f} trained from scratch, and a finetuned XTTS v2. Additionally, we also evaluated the zero-short performance of the pretrained XTTS v2 model and MMS-UNR Mundari model\footnote{\url{https://huggingface.co/facebook/mms-tts-unr}}\footnote{To evaluate the MMS-UNR model, we transliterate our text from Devanagiri to Odia script using \texttt{indic\_nlp\_library}} from Facebook's Massively Multilingual Speech project \cite{pratap2023mms}. Since the data curated is of very high-quality and sampled at 44.1KHz, we trained our VITS models with 44.1KHz data and standard 22.05KHz sub-sampled data. The latter was also used for finetuning the XTTS v2 model. 

Here, we suggest the usage of single E2E models, as they are found to be significantly faster than two-stage models \cite{pmlr-v139-kim21f} and are optimal for deployment and efficient real-time usage. 

\subsection{Training Strategies}
\label{sec:training_strategies}
Both variants of the VITS models were trained with an elevated learning rate of $5e$-$4$ for the generator and discriminator, batch size of 128, and default ExponentialLR scheduler and AdamW \cite{loshchilov2018decoupled} optimizer. As for the XTTS v2 finetuning, a significantly lower learning rate of $5e$-$6$ was used with a batch size of 256, and AdamW with a weight\_decay of $1e$-$2$ was preferred as the optimizer along with a MultiStepLR Scheduler.

All our models were trained, and evaluated on a single A100 80GB GPU and were trained for 2500 epochs and converged within 5 days. A speaker-weighted sampler was also incorporated during the training/finetuning procedure to handle the speaker imbalance on our dataset. The models were checkpointed after every epoch based on the \texttt{loss\_1} of the dev set and the best model checkpoint was used for evaluation.

\section{Results and Discussions}
\label{sec:results_discussions} 

\subsection{Post-Processing}
\label{sec:post-processing}
We use \texttt{ffmpeg} for rudimentary band-pass filtering and noise reduction on synthesized speech. To evaluate the XTTS v2 models, we provide one speaker reference audio from the dev set for conditioning and voice-cloning. Note that, the same reference audio was used for all the test examples for that speaker, and it was manually chosen to be a longer text and speech pair.

\subsection{Subjective Metrics}
\label{sec:sub_metrics}
\begin{figure*}[h!]
    \centering
    \fbox{\includegraphics[scale=0.75]{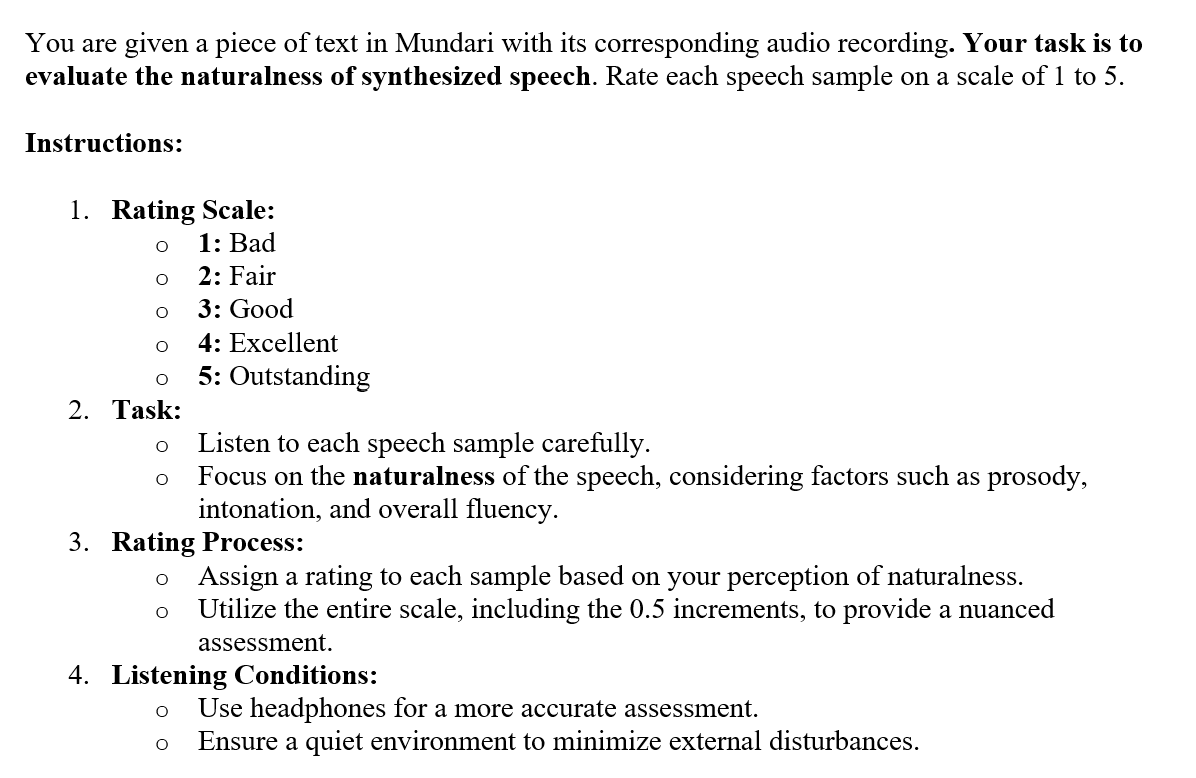}}
    \caption{MOS guidelines provided to the annotators.}
    \label{fig:task_ins}
\end{figure*}

We use the Mean-Opinion-Score (MOS) as the subjective metric for which 100 data points are randomly subsampled from the test set (with speaker stratification). Audio samples generated by various models for this set were sent for human evaluations to native speakers. The task was set up on the Karya platform, and each sample was rated on a scale of 1 to 5 with 0.5 points increments. As discussed earlier, for low-resource languages it is often difficult to find raters for subjective evaluation of the speech samples. In our case, each sample is rated 5 annotators. Using these ratings, we calculate the MOS for the ground truth (both 22.05 KHz and 44.1 KHz) and various models. In total, there were 7 variations for each text sample. Each sample was rated independently, so different variations of a sample were not directly compared. Raters were instructed to use headphones and rate the naturalness of the speech, considering factors such as prosody, intonation, and overall fluency. Detailed instructions are shown in Figure \ref{fig:task_ins}. 
Table \ref{tab:MOS} shows a comparison of the MOS values for the ground truth and the various models. We can see that the VITS-44K model performs the closest to ground truth. We also noticed a huge gap between the VITS model and the other models we studied. Interestingly, the MOS values for XTTS v2 became much worse on finetuning than using it in a zero-shot setup. 

\begin{table}[t!]
\centering
\small
\begin{tabular}{@{}lccc@{}}
\toprule
\textbf{Model} & \begin{tabular}[c]{@{}l@{}}\textbf{Full}\\$n = 100$\end{tabular} & \begin{tabular}[c]{@{}l@{}}\textbf{Male}\\$n = 26$\end{tabular} & \begin{tabular}[c]{@{}l@{}}\textbf{Female}\\$n = 74$\end{tabular} \\ \midrule
\textit{gt-22k}  & 4.62$\pm$0.68 & 4.59$\pm$0.65 & 4.63$\pm$0.69  \\
\textit{gt-44k}  & 4.58$\pm$0.70  & 4.47$\pm$0.79 & 4.62$\pm$0.66 \\ \midrule
\textit{mms}     & 0.79 $\pm$ 1.02 & 0.79 $\pm$ 1.02 & $-$          \\
$\textit{vits-22k}^{\ddag}$ & 3.04 $\pm$ 1.29 & 2.65 $\pm$ 1.34  & 3.18 $\pm$ 1.25  \\
$\textit{vits-44k}^{\dag}$  & 3.69 $\pm$ 1.18 & 3.39 $\pm$ 1.25 & 3.79 $\pm$ 1.13 \\
\textit{xtts-finetuned}  & 0.05 $\pm$ 0.30 & 0.13 $\pm$ 0.52 & 0.02 $\pm$ 0.16 \\
\textit{xtts-pretrained} & 2.20 $\pm$ 1.32  & 2.10 $\pm$ 1.36 & 2.23 $\pm$ 1.31  \\ \bottomrule
\end{tabular}
\caption{MOS values for ground truth and various models. The best and second-best scores are represented by $\dag$ and $\ddag$ respectively. (\textit{gt} = ground truth)}
\label{tab:MOS}
\end{table}

\subsection{Objective Metrics}
\label{sec:obj_metrics}
Mel-Cepstral Distortion (MCD) \cite{407206} is an objective measure used to quantify the difference between two sets of Mel-frequency cepstral coefficients and is useful in evaluating the performance of speech synthesis systems as it provides a numerical indication of how closely the synthesized speech matches the target or reference speech in terms of spectral characteristics. 

For all the models, we compute the MCD scores with dynamic-time wrapping and weighted by speech length with respect to the ground truth subsampled to the sampling rate of the generated speech, if required. We present those in Table \ref{tab:mcd}. Similar to the MOS scores, VITS-44K achieves the lowest error, followed by VITS-22K. Despite XTTS v2 employing speaker conditioning, the scores the significantly worse compared to the best model, VITS-44K. 

XTTS v2 does not natively support Mundari but is pretrained with Hindi, which shares the same characters and pronunciations. Spot-checking some of the audios of the models revealed that the vanilla pretrained XTTS v2 had long pauses between words and made it sound unnatural. However, it captured the intonation and pronunciation well due to its voice-cloning capabilities. The process of finetuning the model resulted in a notable degradation in performance, leading to the generation of nonsensical outputs despite successful convergence. This might due to the catastrophic forgetting induced by the finetuning. We also observed that XTTS v2, which is based on GPT2 \cite{radford2019language}, generated phantom speech in many cases similar to hallucinations in Large Language Models. This phenomenon manifested as the introduction of random Hindi words and gibberish towards the end of the sentence.

\begin{table}[t!]
\centering 
\small
\begin{tabular}{@{}lccc@{}}
\toprule
\textbf{Model} & \begin{tabular}[c]{@{}l@{}}\textbf{Full}\\$n = 1344$\end{tabular} & \begin{tabular}[c]{@{}l@{}}\textbf{Male}\\$n = 350$\end{tabular} & \begin{tabular}[c]{@{}l@{}}\textbf{Female}\\$n = 996$\end{tabular} \\ \midrule 
\textit{mms} & 15.13$\pm$4.19 & 15.13$\pm$4.19 & $-$ \\
$\textit{vits-22k}^{\ddag}$ & 9.45$\pm$3.71 & 10.03$\pm$4.05 & 9.24$\pm$3.56 \\
$\textit{vits-44k}^{\dag}$ & 7.60$\pm$3.99 & 7.27$\pm$3.08 & 7.72$\pm$4.25 \\
\textit{xtts-finetuned} & 13.65$\pm$5.92 & 10.73$\pm$5.33 & 14.69$\pm$5.77 \\
\textit{xtts-pretrained} & 15.80$\pm$7.03 & 13.89$\pm$5.87 & 16.48$\pm$7.27 \\ \bottomrule
\end{tabular}
\caption{MCD scores. The best and second-best scores are represented by $\dag$ and $\ddag$ respectively.}
\label{tab:mcd}
\end{table}

\section{Conclusion}
\label{sec:conclusion}
In this work, we develop a TTS system for a low-resource language, Mundari, a low-resource language spoken by $\approx$ 1M people in India. We also analyze existing models for this language and evaluate popular multilingual and multi-speaker models by finetuning them. We show that the VITS-44K model achieves a mean MOS score of 3.69 and is evaluated as the best among the ones compared by native speakers. We release our model publicly and hope this research further promotes the development of speech systems for endangered and low-resource languages, aiding in bridging the digital divide in India.
\section{Limitations}
\label{sec:limitations}

\begin{itemize}
    \item Our primary emphasis in this study centers on E2E TTS, deliberately excluding the consideration of combinations involving Acoustic models and Vocoders, as observed in prior works \cite{vakyansh,prakash20_interspeech,10096069}. The motivation behind this choice is the intention to construct a simple unified system for speech synthesis, designed for straightforward deployment and ease of use by the general public. 
    \item We explicitly recognize the inherent bias in the speaker distribution employed for our study. The challenge of recruiting native proficient speakers, capable of dedicating extended hours and effort to the recording process, contributed to a noticeable synthesis disparity, particularly evident in the diminished quality of male speech synthesis outputs.
\end{itemize}
\section{Ethical Considerations}
\label{sec:ethics}
We use the framework by \citet{bender-friedman-2018-data} to discuss the ethical considerations for our work.

\begin{itemize}
    \item \textbf{Institutional Review:} All aspects of this research were reviewed and approved by Karya.
    \item \textbf{Data:} Our data is collected in multiple steps as described in section \ref{sec:data}. We first source the Hindi sentences and manually translate them to Mundari. Specific guidelines for translations were provided. These Mundari sentences were then recorded in a studio by 2 speakers.
    \item \textbf{Speaker Demographic:} We recruited 2 speakers to record the audio. Their payment was set after deliberation and contracts were signed. Speakers were paid INR 8 per recording. The average duration of a sample is $\approx$ 3.7 seconds. 
    \item \textbf{Annotator Demographics:} Annotators for MOS rating were recruited through an external annotator services company.
    All annotators were native speakers of the language. The pay was INR 2 per sample, with an average sample length of $\approx$ 3.7 seconds. 
    \item \textbf{Annotation Guidelines:} We draw inspiration from the community standards set for similar tasks. These guidelines were created following best practices after careful research. Annotators were asked to rate the speech samples on naturalness. A detailed explanation was given for the task. Annotator's identity was hidden from the authors to limit any bias. 
    \item \textbf{Methods:} In this study, using our Mundari speech dataset, we trained 2 models: VITS-22K and VITS-44K, and finetuned the XTTS v2 model. We release the models for the benefit of the Mundari and the research community.
\end{itemize}
\section{Acknowledgements}
We sincerely thank the voice artists, Roshan and Meenakshi, for lending us their voices to create the speech dataset. We also extend our gratitude to Prof. Bornini Lahiri and Prof. Dripta Piplai from IIT Kharagpur for their advice on data collection and processing. Finally, we thank Praveen SV (Ph.D. Student, IIT Madras) and Gokul Karthik (MLE, Technology Innovation Institute) for their walk-throughs of IndicTTS\footnote{\url{https://github.com/AI4Bharat/Indic-TTS}} and Coqui.

\bibliography{anthology,custom}

\end{document}